\documentclass[10pt,twocolumn]{article}
\usepackage{amsmath,amssymb}
\usepackage{graphicx}
\newcommand{\tmem}[1]{{\em #1\/}}
\newcommand{\tmop}[1]{\operatorname{#1}}

\newtheorem{definition}{Definition}
\newtheorem{theorem}{Theorem}
\def\Pr{\hbox{\sf Pr}}

\begin{document}

\author{Thomas M. Breuel\\PARC, Palo Alto, USA\\{\tt tmb@parc.com}}

\date{November 2003\footnote{This paper was originally written
in November 2003, but has been submitted to Arxiv in 2007.
References have not been updated to include more recent work.}
}

\title{\Large\bf Learning View Generalization Functions}

\maketitle

\section*{\centering Abstract}
{\it Learning object models from views in 3D visual object recognition is
usually formulated either as a function approximation problem of a function
describing the view-manifold of an object, or as that of learning a
class-conditional density. This paper describes an alternative
framework for learning in visual object recognition, that of learning the
view-generalization function. Using the view-generalization function, an
observer can perform Bayes-optimal 3D object recognition given one or more
2D training views directly, without the need for a separate model
acquisition step. The paper shows that view generalization functions can be
computationally practical by restating two widely-used methods, the
eigenspace and linear combination of views approaches, in a view
generalization framework. The paper relates the approach to recent methods
for object recognition based on non-uniform blurring. The paper presents
results   both on simulated 3D ``paperclip'' objects and real-world images from the
COIL-100 database showing that useful view-generalization functions can be
realistically be learned from a comparatively small number of training
examples.}

\section{Introduction}

Learning view-based or appearance-based models of objects has been a
major area of research in visual object recognition (see
{\cite{Edelman97}} for reviews).  One direction of research has
focused on treating the problem of learning appearance based models as
an {\tmem{interpolation problem}} {\cite{UllBas91, PogEde90}}.
Another approach is to treat the problem of learning object models as
a \textit{classification problem}.

Both approaches have some limitations.  For example, acquiring a novel
object may involve fairly complex computations or model building.
They also do not easily explain how an observer can transfer his skill
at recognizing existing objects to generalizing from single or
multiple views of novel objects; to explain such transfer, a variety
of additional methods have been explored in the literature, including
the use of object classes or categories, the acquisition and use of
object parts, or the adaptation and sharing of features or feature
hierarchies.

This paper describes an approach to learning appearance-based models
that addresses these issues in a unified framework: the visual
learning problem is reformulated as that of learning \tmem{view
generalization functions}.  The paper shows that knowledge of the view
generalization function is equivalent to being able to carry out
Bayes-optimal 3D optimal object recognition for an arbitrary
collection of objects, presented to the system as training views.
Model acquisition reduces to storing 2D views and does not involve
learning or model building.

This represents a significant paradigm shift relative to previous
approaches to learning in visual object recognition, which have
treated the problem of acquiring models as a separate learning
problems.  While previous models of visual object recognition can be
reinterpreted in the framework in this paper (and we will do so for
two such methods), the formulation in terms of view generalization
functions makes it easy to apply any of a wide variety of standard
statistical models and classifiers to the problem of generalization to
novel objects.

In this paper, I will first express Bayes-optimal 3D object
recognition in terms of training and target views and prior
distributions on object models and viewpoints.  Then, I will describe
the statistical basis of learning view generalization functions.
Finally, I will demonstrate, both on the standard ``paperclip'' model
and on the COIL-100 database, that learning view generalization
functions is feasible.

\section{Bayesian 3D Object Recognition}

This section will review 3D object recognition from a Bayesian
perspective and establish notation.
Let us look at the question of how an observer can recognize 3D
objects from their 2D views.  Let $\omega$ identify an object and $B$
be an unknown 2D view (we will refer to $B$ also as the {\em target
view}).  Then, classifying $B$ according to $\hat{\omega}(B) =
\tmop{arg} \max_{\omega} P ( \omega | B )$ is well known to result in
minimum error classification \cite{Duda01}.  Using Bayes rule, we can
rewrite this as
\begin{eqnarray}\label{bayesrule}
  \tmop{arg} \max_{\omega} P ( \omega | B ) & = & \tmop{arg} \max_{\omega} 
  \frac{P ( B | \omega ) P ( \omega )}{P ( B )}\\
  & = & \tmop{arg} \max_{\omega} P ( B | \omega ) P ( \omega ) \nonumber
\end{eqnarray}
$P(\omega)$ is simply the frequency with which object $\omega$
occurs in the world.  Let us try to express $P ( B | \omega )$
in terms of models and/or training views.

Assume that we are given a 3D object model $M_{\omega}$.  In the absence
of noise, the projection of this 3D model into a 2D image is determined by
some  function $f$ of the viewing parameters $\phi \in \Phi$, $B = f (
M_{\omega}, \phi )$.  The function $f$ usually is rigid body transformations
followed by orthographic or perspective projection. 

In the presence of additive noise, $B = f ( M_{\omega}, \phi ) + N$
for some amount of noise distributed according to some prior noise
distribution $P ( N )$.  With this notation, we can now express $P ( B
| \omega )$ in terms of the 3D object model\footnote{$\delta$ is
the Dirac delta function.}
\begin{equation}\label{objviews}
P ( B | \omega ) = \int \delta ( B, f ( M_{\omega}, \phi ) + N ) P ( \phi )
   P ( N ) \; d \phi \; d N
\end{equation}
To simplify notation below, we write
$P(B|M_\omega,\phi)=\int \delta ( B, f ( M_{\omega}, \phi ) + N )\,P(N)\,dN$
and obtain
\begin{eqnarray}\label{mbr}
  P ( B | \omega ) = \int P ( B | M_{\omega}, \phi ) P ( \phi ) d \phi &  & 
\end{eqnarray}
By construction, Equation~\ref{mbr} represents {\em Bayes-optimal 3D
model-based recognition}, assuming perfect knowledge of the 3D model
$M_\omega$ for a given object $\omega$.

\begin{figure}[t]
\centerline{\includegraphics[height=0.7in]{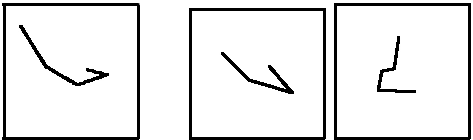}}
\caption{Examples of paperclips used in the simulations.}
\label{figexclip}
\label{figexamples}
\end{figure}

In real-world recognition problems, the observer is rarely given a
correct 3D model $M_{\omega}$ prior to recognition.  Instead, the
observer needs to infer the model from a set of training views\footnote{
For the rest of the paper, we limit ourselves to
the case where the training and test views are drawn in an identical
manner and independently of one another; the more general case in
which, say, the training views $\mathcal{T}_\omega$ come from a motion
sequence and hence have sequential correlations in their viewing parameters
can be treated analogously.
}
$\mathcal{T}_{\omega} = \{ T_{\omega,1}, \ldots, T_{\omega,r} \}$.
Therefore, an observer is faced with the problem of determining $P ( B |
\omega )$ as $P ( B | \mathcal{T}_{\omega} )$.  In a model-based framework,
this means that the observer attempts to perform reconstruction of the object model $M$
given the training views $\mathcal{T}_{\omega}$ and then performs
recognition using the resulting distribution of probabilities over the
possible models for recognition.  If we put this together with Equation~\ref{mbr},
we obtain for $P(B|\omega) = P(B|\mathcal{T}_\omega)$:
\begin{equation}\label{viewgenstat}
  P ( B | \mathcal{T}_{\omega} ) = \int P ( B | M, \phi ) P ( M | \mathcal{T}_{\omega} ) P ( \phi ) d M d \phi
\end{equation}
By construction, $P(B|\mathcal{T}_\omega)$ represents the density of
target views $B$ given a set of training views $\mathcal{T}_\omega$.
Therefore, applying Equation~\ref{viewgenstat} together with
Equation~\ref{bayesrule} results in {\em Bayes-optimal 3D model-based
recognition from 2D training views}.

Now that we have derived the Bayes-optimal 3D object
recognition, let us look at some approaches that have been proposed in
the literature for solving the 3D object recognition problem and how
they relate to Bayes optimal recognition.

\begin{figure}[t]
\includegraphics[height=1in]{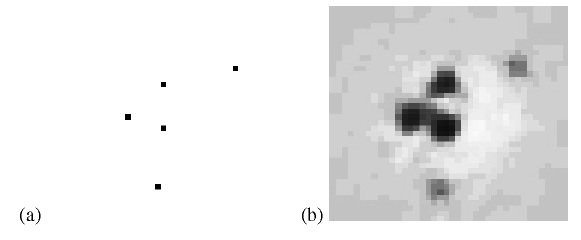}
\caption{
Illustration of $P(B|T_\omega)$.  (a) The feature vector $T_\omega$,
represented as an image (vertices of the clip quantized to a grid),
(b) $\log \hat{P}(B|T_\omega) - \log \hat{P}(B)$ (darker=higher
probability).
\label{figpost}
}
\end{figure}

\paragraph{3D Model-Based Maximum Likelihood Methods.}

Traditional approaches to model-based 3D computer vision (e.g.,
\cite{Grimson90z}) generally divide recognition into two phases.
During a model acquisition phase, the recognition system attempts to
optimally reconstruct 3D models from 2D training data.  During the
recognition phase, the system attempts to find the optimal match of
the reconstructed 3D model against image data.

This is often realized by estimating $M_{\omega}$ using a maximum
likelihood or maximum a posteriori (MAP) procedure (e.g., least square
methods, assuming Gaussian error), $\hat{M}_{\omega} = \tmop{arg}
\max_M P ( M | \mathcal{T}_{\omega} )$ and then performing 3D
model-based recognition in a maximum likelihood setting using
$\hat{M}_{\omega}$.
\begin{eqnarray}\label{ccv-mlv}
  P ( B | \omega ) &=& P ( B | \mathcal{T}_{\omega} ) = \max_{\phi} P ( B |
  \hat{M}, \phi ) \\
\hat{M} &=& \tmop{arg} \max_M P ( M |
  \mathcal{T}_{\omega} ) \label{ccv-mlm}
\end{eqnarray}

It is important to remember that this approach is not Bayes optimal in
general--it is a good approximation only under certain conditions, for
example, when all the distributions $P ( B | M, \phi )$ are unimodal,
sharply peaked, and have comparable covariances.  Furthermore,
computationally, the maximum likelihood estimations have proven to be
fairly difficult and costly optimization problems.

One reason that has made such approaches attractive is that, as the
amount of noise and variability become small, the reconstruction
and matching problems can be treated geometrically, and a wealth
of results has been derived in that limit (c.f. algorithms like
\cite{higgins81}).  But from a statistical point of view, such geometric
approaches can be unnecessarily restrictive. For example, in the case
in which the training set $\mathcal{T}_{\omega}$ consists of only a
single view $T_\omega$, 3D reconstruction is not possible for
arbitrary 3D objects.  Yet, as we will see in the experimental results
below, $P ( M | T_\omega )$ still contains considerable amounts of
information.

\paragraph{View Interpolation Approaches.}

Because the imaging transformation $f ( M, \phi
)$ is smooth, the set of views $\mathcal{B}_M = \{ f ( M, \phi ) | \phi \in
\Phi \}$ of an object itself forms a smooth, low-dimensional surface in the
space of all possible views.  In fact, $\mathcal{B}_M$ is embedded in a
low-dimensional linear subspace of the space of all possible views
{\cite{UllBas91}}.  The smoothness of $\mathcal{B}_M$ suggests that it might
be learned from examples using a surface or function interpolation method. 
This has given rise to one of the most influential approaches to learning in
3D object recognition, developed by Poggio and Edelman {\cite{PogEde90}}.

Methods that approximate the view manifold (e.g.,
\cite{PogEde90,UllBas91,nayar96realtime}) generally attempt to compute
some geometrically motivated distance of the target view from the view
manifold and then perform nearest neighbor classification in terms of
that distance.  This approach would minimize recognition error rates
if the distribution of views over the view manifolds were uniform and
several other conditions were satisfied.  However, most work on
geometric and interpolation methods does not demonstrate
Bayes-optimality of the classification error, but only proves results
about the quality of the approximation to the view manifold that they
achieve.  In general, a good approximation to the view manifolds is
neither necessary nor sufficient for Bayes-optimal recognition
(although it does often seem to work reasonably well).

\paragraph{Classification Approaches.}

Many classification methods (multi-layer perceptron, logistic
regression, mixture discriminant analysis, etc.) are concerned with
estimating posterior distributions like $P ( \omega | B )$ or
corresponding discriminant functions directly.  They share with the
methods described in this paper that they do not necessarily involve
the two-step maximization procedure used in traditional model-based
systems (Equations~\ref{ccv-mlv} and~\ref{ccv-mlm}).  Classification
methods have not been all that popular for 3D object recognition in
the past, but there has been some recent work on it (e.g.,
\cite{pontil98support}).

\paragraph{Single-View Generalization.}

Based on geometric considerations alone,
if nothing else is known about a 3D object, multiple views of an
object are needed in order to reconstruct a 3D model of the object
from views (e.g., \cite{higgins81}).  Generalization from a single
view is usually only considered possible when the object is known to
have special properties like symmetry or when the object is known to
be a member of some other kind of object class (e.g.,
\cite{vetter97linear}).  Geometrically, of course, this is true.
Statistically, however, even if 3D model reconstruction is not
possible, $P(B|\mathcal{T}_\omega)$ may still contain information
permitting significant single view generalization, as the experiments
below will show.

\section{View Generalization Functions}

We have seen that previous approaches to learning object models have
concentrated on learning $f_M ( \omega )$, $P ( \omega | B )$, or $P (
B | \omega )$.  This paper proposes and examines a different learning
problem for 3D object recognition: the direct estimation of the view
generalization function, defined as follows:

\begin{definition}
We define the {\bf $r$-view generalization function}
as the conditional density
$P(B|\mathcal{T}_{\omega}) = P( B | T_{\omega,1}, \ldots,
T_{\omega,r})$ given by Equation~\ref{viewgenstat}.
\end{definition}

If the training set $\mathcal{T}_\omega$ consists of a single view
$T_\omega$, we call this a {\em single view generalization function}.
Notice that view generalization functions are functions of views only;
they do not involve any object models.  In some sense, they tell us
how much an unknown view is similar to a set of training views. 

If we have a good estimate of the view generalization function, we can
perform Bayes-optimal 3D object recognition by a generalized nearest
neighbor procedure with a variable metric, somewhat analogous to the
procedure in \cite{Lowe95}.

That is, the vision system initially builds a good approximation of
the view generalization function ${P}(B|\mathcal{T}_\omega)$ from
visual input.  This might require a lot of training data,
corresponding perhaps to several years of visual input after birth in
human vision.

Once a vision system has acquired a fairly good approximation of
${P}(B|\mathcal{T}_\omega)$, the acquisition of new object models merely
required storing the training views $\mathcal{T}_\omega$.  Let us
assume that training views are unambiguous, $P(\omega|T_\omega)=1$
(otherwise, the procedure is still optimal $k$-nearest neighbor but
does not necessarily achieve Bayes-optimal classification rates
\cite{2003-breuel-icdar-2}).  Given the view generalization function
and a collection of training views for each object, Bayes-optimal
recognition of an unknown view $B$ against the model base can then be
carried out by evaluating ${P}(B|\mathcal{T}_{\omega_i})\,P(\omega_i)$
for each object $\omega_i$ under consideration and classify according
to Equation~\ref{bayesrule}.  Furthermore, if the view generalization
function ${P}(B|\mathcal{T}_\omega)$ can be implemented in a low-depth
circuit, the visual system will be able to carry out Bayes-optimal
recognition of novel 3D objects from 2D training views quickly,
without the need for the optimizations implicit in traditional maximum
likelihood approaches used in computer vision (see Equations~\ref{ccv-mlv}
and~\ref{ccv-mlm}).

Of course, whether this approach works hinges crucially on whether it
is possible to learn an approximation to the view generalization
function that actually generalizes to novel objects and has the
desired properties.  If every new object the system encounters
requires updating of the estimate of the view generalization function
and the approach effectively reduces to traditional one-by-one
learning of object models.  If, on the other hand, after an initial
set of training examples, the estimate of ${P}(B|\mathcal{T}_\omega)$
generalizes reasonably well to previously unseen objects, then the
approach is successful.

The rest of this section will explore these issues further with
examples and some theoretical arguments.  Subsequent sections will
provide some experimental evidence that learning view generalization
functions is feasible.

\paragraph{Smoothness of the View Generalization Function.}

Intuitively, we would expect that, for most objects and views, if the
set of training views $\mathcal{T}_\omega$ for two objects is similar,
the distributions $P(M|\mathcal{T}_\omega)$ of possible corresponding
object models are similar as well, and so are the distributions
$P(B|M)$ of other possible views.  This corresponds to a statement
about the smoothness of the view generalization function.  It can be
demonstrated formally for specific model distributions, camera and
noise models by differentiating Equation~\ref{viewgenstat} with
respect to $B$ and the $T_{\omega,i}$.

Such smoothness properties suggest that the view generalization
function may be learnable using techniques like radial basis function
(RBF) interpolation or multilayer perceptrons (MLPs) that take
advantage of smoothness; \cite{PogEde90} use a similar argument to
motivate the use of RBFs for learning individual view manifolds.

Note that, in contrast to the view generalization function, the
maximum likelihood solutions given by Equations~\ref{ccv-mlv}
and~\ref{ccv-mlm} and used in many computer vision systems, when
viewed as functions of the target and training views, are not
necessarily smooth and therefore probably not easily approximated
using models like RBFs.

\paragraph{Model Priors.}

One of the important properties of the view generalization function is
that it does not depend on the specific models the observer has
acquired in his model base.  Rather, it depends on the prior
distribution of models from which the actual models encountered by the
system are drawn.

\begin{theorem}
The view generalization function is completely determined by the prior
distribution of 3D models $P(M)$, the distribution of viewing
parameters $P(\phi)$, the noise distribution $P(N)$, and the choice of
imaging model $f(M,\phi)$.
\end{theorem}

{\it Proof.}  In analogy to Equation~\ref{objviews}, we have for a
training view $T_\omega$, $P(T_\omega|M) =
\int\delta(T_\omega|f(M,\phi)+N)\,P(\phi)\,P(N)\,d\phi\,dN$.
Since the training views are (by assumption) drawn independently,
$P(\mathcal{T}_\omega|M) = \prod_{T_\omega\in\mathcal{T}_\omega}P(T_\omega|M)$.
Using Bayes formula, we invert this to yield $P(M|\mathcal{T}_\omega)$.
Furthermore, $P(B|M,\phi) = \delta(T_\omega|f(M,\phi)+N)\,P(N)\,d\phi\,dN$.
With this, we have all the components to evaluate Equation~\ref{viewgenstat}.
$\Box$

\paragraph{Linear Combination of Views.}

Let us now turn to the question of whether fast, or even low-depth
arithmetic circuit, implementations of view generalization functions
are plausible.  To do this, we will recast two commonly used
approaches to 3D object recognition, linear combination of views
\cite{UllBas91} and eigenspace methods (below), into a
view-generalization function form.  The resulting view generalization
functions implement those models exactly and hence would perform
identically to those methods if implemented.

In a linear combination of views framework, we test whether a novel
target view $B$ can be expressed as a linear combination of training
views.  Let us assume concretely that we want to generalize based on
three training views per object, $P ( B | T_1, T_2, T_3 ) = g ( B,
T_1, T_2, T_3 )$. The error $\epsilon$ by which we judge similarity is
the magnitude of the residual that remains after the linear
combination of training views has been subtracted. Performing nearest
neighbor classification using $\epsilon$ corresponds to assuming any
of a wide number of unimodal, symmetric distributions $U$ for
$\epsilon$; that is, nearest neighbor classification using linear
combination of views is the same as classifying using the conditional
density $P ( B | T_1, T_2, T_3 ) = U ( \epsilon^{} )$.  If we write
$\rho_v ( x ) = x - \frac{v \cdot x}{\| v \|} v$ for the residual that
remains after subtracting the projection of $x$ onto $v$ from $x$,
then we can compute $\epsilon$ as $\epsilon = \| \rho_{T_3} (
\rho_{T_2} (
\rho_{T_1} ( B ))) \|$, and the linear combination of views (LCV) view
generalization function $g_{\tmop{LCV}} ( B, T_1, T_2, T_3 ) = U ( \epsilon )
= U ( \| \rho_{T_3} ( \rho_{T_2} ( \rho_{T_1} ( B ))) \| )$.  Generalizing to
$r$ training views, we can clearly compute this with an arithmetic circuit of
depth proportional to $r$. Therefore, we have seen that if we use a linear
combination of view model of object similarity, then the view generalization
function can be expressed as a fairly simple function that can be implemented
as a circuit of depth proportional to the number of views $r$.

\paragraph{Eigenspace Methods.}

Eigenspace methods and related techniques have been used extensively
in information retrieval (latent semantic analysis, LSA) and computer
vision {\cite{moghaddam_cvpr94,nene96columbia}}.  In general, in
eigenspace methods, given a set of training views $T_i$ for multiple
objects, we compute a low-dimensional linear subspace $\mathcal{S}$
and evaluate similarity among a target view $B$ and a training view
$T_{\omega}$ within that low-dimensional subspace.  That is,
eigenspace methods use an error $\epsilon = \| \Pr_{\mathcal{S}} ( B )
-
\Pr_{\mathcal{S}} ( B ) \|$ for nearest neighbor classification, where $\Pr_\mathcal{S}$ is
the linear projection operator onto $\mathcal{S}$.  This
procedure can be justified, for example, when the training samples $T_i$
falls into a low-dimensional linear subspace in the error free
case, but are corrupted with Gaussian noise whose magnitude is small
compared to the variability of the training samples.  Then, if we
determine the covariance matrix of the $T_i$, its large eigenvalues
will correspond approximately to directions representing meaningful
object variability, while its small eigenvalues will correspond
approximately to directions representing only noise \cite{Duda01}.

As before, nearest neighbor classification using $\epsilon$ is
equivalent to choosing some unimodal error distribution $U ( \epsilon
)$ (e.g., Gaussian) and approximating
\begin{equation}
  P ( B | \mathcal{T}_{\omega} ) \propto \max_{T \in \mathcal{T}_{\omega}} U (
  \epsilon ) = \max_{T_{} \in \mathcal{T}_{\omega}} U ( \| P_{\mathcal{S}} ( B
  ) - P_{\mathcal{S}} ( B ) \| ) \label{esvg}
\end{equation}
Therefore, we can view eigenspace methods as a very simple form of learning a
view generalization function; the function has the specific form given in
Equation \ref{esvg}, with only the projection operator $\Pr_{\mathcal{S}}$ being
learned by the observer.

\section{First Order Single View Model}

In this section, we will look at a simple experimental evaluation of
single view generalization functions, applied to simulated 3D
paperclips.  Simulated 3D paperclips are widely used in computational
vision, psychophysical experiments, and neurophysiological work (e.g.,
\cite{PogEde90,LiuKniKer95}).  Let us briefly review the model here
and state the parameters used in this and the next section.

Random 3D models are generated by picking five unit vectors in
$\mathbb{R}^3$ with uniformly random directions and putting them
end-to-end. To obtain a 2D view of the object, the 3D model is rotated
by some amount and then projected orthographically along the $z$
axis. Views are centered so that the centroid falls at the origin.

For all the experiments involving paperclips below, the training set
consisted of random views derived from a fixed set of 200 randomly
constructed 3D clip models. That is, all generalization to arbitrary,
previously unseen 3D clip models was derived from information learned
from this small, fixed sample of 200 clips.

For each test trial, novel previously unseen 3D clip models were
generated randomly and random views of those clips were generated by
random rotations in the range $[-40^{\circ},+40^{\circ} ]$ around the
$x$ and $y$ axes relative to the training view; this range of
rotations was chosen because it is comparable to what previous
authors have used and seems to be at the limit of human single view
generalization ability for these kinds of images (e.g.,
\cite{PogEde90}).

In order to be accessible to a learning algorithm, these views need to
be encoded as a feature vector. Three kinds of encodings have been
commonly used in the literature and are used in this paper. An angular
encoding uses the ordered sequence of angles around each vertex in the
projected image, giving rise to a four-dimensional feature vector.  An
ordered location encoding uses the concatenation of $x$ and $y$
coordinates, in sequence, as its feature vector, resulting in a 10
dimensional feature vector.  A feature map encoding projects the
vertices of the clip onto a bounded grid composed of $40\times40$
buckets, resulting in a binary feature vector of length $1600$.

\paragraph{Single View Generalization.}

Let us now look at building an empirical distribution model of $P ( B
| \mathcal{T}_{\omega} )$.  We will limit ourselves to
{\tmem{single-view generalization models}}; that is, we assume that
the set of training views for an object $\omega$ consists of a single
view $\mathcal{T}_{\omega} = \{ T_{\omega} \}$.  Note that this
problem has not been studied much in computer vision; this is perhaps
because, based on geometry alone, a training set consisting of a
single view $T_{\omega}$ does not permit reconstruction of the 3D
structure of an arbitrary object even in the error-free case. However,
as several authors have observed (e.g., {\cite{PogEde90}}), human
observers are capable of a significant degree of 3D generalization, so
there is reason to believe that 3D recognition based on $P ( B |
T_{\omega} )$, that is, recognition based solely on a single training
view is possible, at least to some degree.

\paragraph{First Order Approximation.}

For concreteness, let us assume the feature map representation of
views discussed above. In that representation, a view $B$ is a binary
feature vector $B = ( B_1, \ldots, B_r )$, where each $B_i$ represents
a pixel or bucket in the image, and analogously for $T$.  We can try
to model $P ( B | T )$ as a an expansion {\cite{mccullagh87}}:

\begin{equation}
\log P ( B | T ) \approx
     \frac{1}{Z} ( h^{( 0 )} + \sum_{i j} h_{i j}^{( 1)} ( B_i, T_j ) 
      + \sum_{i j k} h^{( 2 )}_{i j k} ( B_i, T_j, T_k ) + \ldots )
\end{equation}
Here, the $h^{( k )}$ are functions of their boolean-valued arguments. The
different $h^{( k )}$ correspond to taking account increasingly higher-order
correlations among features.

Of particular interest is the ``first-order'' approximation, for which we take
into account only $h^{( 0 )}$ and $h^{( 1 )}$.  Let us look at the probability
that pixel $B_i$ in the view $B$ is ``on'' given the training view $T$:
\[ \log P ( B_i = 1 | T ) \propto \tmop{const} + \sum_{i j} h_{i j} ( 1, T_j )
\]
But this means that if we look at $\log P ( B_i | T )$, it is a
blurred version of the training view, with with $h_{ij}$ as a
spatially varying blurring kernel.

Blurring, with or without spatially variable kernels, has been proposed as a
means of generalization in computer vision by a number of previous authors. 
In a recent result, {\cite{berg01}} derives non-uniform blurring for 2D
geometric matching problems, the ``geometric blur'' of an object.   The
results sketched in this section make the connection between non-uniform
geometric blurring and first order approximations to the single view
generalization function, $g ( B, T ) = P ( B | T )$.  This connection lets us
determine more precisely how we should compute geometric blurring, what
approximations it involves compared to the Bayes-optimal solution, and how we
can improve those approximations to higher-order statistical models.  Let us
note also that there is nothing special about the representation in terms of
feature maps; had we chosen to represent views as collections of feature
coordinates, a first order approximation would have turned into error
distributions on the location of each model feature.

\paragraph{Experimental Results.}

Using the paperclip models, we can estimate the
parameters of the first order model above by simulation: we repeatedly
generate different views of objects, compute their feature vectors, and
compute the frequency of co-occurrence of features in the training view $T$
and a target view $B$ (a kind of Hebbian learning).  This allows us to
visualize the non-linear blurring that results in single-view generalization. 
An example of this is shown in Figure \ref{figpost}.

Note that, similar to
\cite{berg01}, there is more blurring further away from the center of the
object.  However, the two approaches differ in that geometric blur
does not take into account, among other things, the prior distribution
of models $P(M)$ and hence does not necessarily result in Bayes
optimal performance when applied to object recognition problems, while
the empirical statistical model of view similarity used here
approximates the true class conditional distribution.

In terms of error rates in a forced choice experiments, view
similarity using these non-uniform blurs achieves an error rate of
7.2\%, compared to 32\% using simple 2D similarity, demonstrating
substantial improvements from the use of the view similarity approach.
Note also that because of the nature of the feature vector used--a 2D
feature map--the system did not have access to correspondence
information.

\section{View Similarity Models}

Densities like the view generalization function $P ( B |
\mathcal{T}_{\omega} )$ can be difficult to estimate.
It would be more convenient if we could reformulate the learning
problem as that of modeling a class posterior density: there is a
wide variety of models available for class posterior density
(logistic regression, radial basis functions, multilayer-perceptrons,
etc.)

Fortunately, we can perform that transformation fairly easily.  During
recognition from a model base, we compare the unknown view $B$ repeatedly
against collections of training views $\mathcal{T}_{\omega}$ for each object. 
There are two conditions under which this takes place: either the view $B$
derives from the same object $\omega$ as the training views
$\mathcal{T}_{\omega}$, or the view derives from some other object.  Let us
represent these two conditions by a boolean indicator variable $S$.  For $B$
not derived from $\omega$, the conditional distribution $P ( B | S = 0,
\mathcal{T}_{\omega} )$ is simply the prior distribution of possible views $P
( B )$.  When $B$ is derived from the same object as the training views, that
is $S = 1$, we have:
\begin{eqnarray*}
  P ( B | S = 1, \mathcal{T}_{\omega} ) & = P ( S = 1 | B,
  \mathcal{T}_{\omega} ) & \frac{P ( B )}{P ( S = 1 | \mathcal{T}_\omega )}
\end{eqnarray*}
Given an unknown view $B$ to recognize, $P ( B )$ does 
not change with $\omega$, and $P(S=1|\mathcal{T}_\omega)=P(\omega)$.
Therefore,
\[ \hat{\omega} = \tmop{arg} \max_{\omega} P ( B | \mathcal{T}_{\omega} ) 
		P(\omega) =
   \tmop{arg} \max_{\omega} P ( S = 1 | B, \mathcal{T}_{\omega} ) 
   \]
Let us call the distribution $P ( S = 1 | B, \mathcal{T}_{\omega} )$ the
{\tmem{view similarity function}}.  If $\mathcal{T}_{\omega}$ consists of a
single view, we call this distribution the {\tmem{single view similarity
function}}.  It acts like an adaptive similarity metric \cite{Lowe95} when
used for recognition from a model base using Equation~\ref{bayesrule}.

\begin{figure}[t]
\centerline{(a)~\includegraphics[height=0.7in]{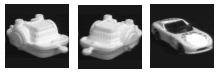}
(b)\includegraphics[height=0.7in]{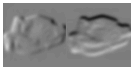}}
\caption{(a) Sample images from the COIL-100 database.
(b) The feature map used as input to the recognition system.}
\label{figcoilex}
\end{figure}

\paragraph{Experiments.}

Let us look now at how view similarity functions can be learned in an
the case of 3D paperclips.  As in the previous section, we consider
the single view generalization problem and apply it to the problem of
paperclip recognition.  During a training phase, the experiments used
a collection of 200 paperclips, generated according to the procedure
described in the previous section. The procedure used for generating
the paperclips implies the prior distribution $P ( B ) = P (
T_{\omega} )$, and the training set is a sample from this
distribution. For training, the system chooses one of those paperclips
$\omega$ at random and generates two different views, a training view
$T_{\omega}$, and a target view $B$.  Then, it picks a second
paperclip $\omega' \neq \omega$ at random and generates a view $B'$.
The pair $( B, T_{\omega} )$ is then a training example for the
condition $S = 1$, and the pair $( B', T_{\omega} )$ is a training
example for the condition $S = 0$. Generating a number of these pairs,
we obtain a training set for a Bayesian classifier $\tilde{P} ( S | B,
T_{} )$.

For testing, the experiment was carried out using novel
paperclips--paperclips not found in the training set of 200
paperclips.  We could test by generating a model base of some number
of objects and then performing nearest neighbor classification; we
will do that below on the COIL-100 database of real images. However, that
introduces another unnecessary parameter into the evaluation, the size
of the model base.  Therefore, here, we reduce the recognition
problems on a forced choice experiment. In such a forced-choice
experiment, we generate test samples analogous to training samples and
measure the error rate of the system on being able to distinguish $(
B, T_{\omega} )$ from $( B', T_{\omega} )$.  This is also a common
paradigm used in psychophysical experiments.  An example of such a
forced choice experiment can be seen in Figure~\ref{figexamples}; the
image at the left is the training view $T_\omega$, and the two images
on the right correspond to $B$ and $B'$ (not necessarily in that
order).  Views were encoded using the three feature types described
in the previous section; for location features, rotations were
chosen from $\{\pm 45^\circ\}$.

  \begin{center}
    \begin{tabular}{|c|c|c|c|}
      \hline
      & \multicolumn{3}{c|}{Error Rate} \\
      \hline
      & Angles & Locations & Feature Map\\
      \hline
      2D Similarity & 19.9\% & 8.4\% & 32\% \\
      \hline
      View Similarity & 10.9\% & 0.38\% & 7.9\% \\
      \hline
    \end{tabular}
  \end{center}

These results show a substantial improvement of view-similarity
functions over 2D similarity on single view generalization to novel
objects.  Note that many traditional recognition methods, like linear
combinations of views or model-based recognition, cannot even be
applied to this case because the observer is only given a single
training view for each novel object.

\section{Experiments with COIL-100}

The experiments in the previous sections were all carried out on
simulated 3D paperclip objects--a widely used test case in the
literature.  However, real-world images might show considerably more
variation and hence make the learning of view generalization functions
hard or impossible from reasonable numbers of training images.

To test whether view similarity methods are applicable to real images,
experiments were carried out on the COIL-100 database
\cite{nene96columbia}.  Furthermore, the eigenspace method used in
\cite{nayar96realtime} was implemented as a control.

The COIL-100 database contains color images representing views of
objects separated by $5^\circ$ rotation around the vertical axis.
Even simple nearest neighbor classification methods perform nearly
perfectly given that sampling and color input, so using the full
database as training examples is not a very hard test of the
ability to generalize to new views based on shape.

To test for the ability to generalize to viewpoints that differ
substantially from the training view based on shape alone, the
database was preprocessed to remove color and absolute intensity
information, and only a coarser sampling of viewpoints was used.
Images were converted to grayscale and gradient features were
extracted, as shown in Figure~\ref{figcoilex}.  Training was carried
out on views from the first 70 objects in the database.  The methods
were tested on views from the remaining 30 objects of the database.
For each test, only collections of views whose viewpoints were spaced
apart by multiples of $30^\circ$ (12 per object) were used.

The question addressed by these experiments on the COIL-100 database
is whether it is possible to learn view generalization functions that
are capable of any kind of generalization at all.  Note that the view
similarity model had no prior knowledge incorporated into it at all,
not even Euclidean distance.  Without effective training, the view
similarity function performs at chance level, an error rate of 96.7\%.
Any performance better than that means that the view similarity model
successfully generalized at least to some degree from the 70 training
objects to the 30 previously unseen test objects.  Error rates for
this recognition problem are shown in the following table (measured
for 2160 test views):

  \begin{center}
    \begin{tabular}{|c|c|}
      \hline
      & Error Rate\\
      \hline
      Euclidean Distance & 40.0\% \\
      \hline
      Eigenspace & 26.1\% \\
      \hline
      View Similarity & 20.3\% \\
      \hline
    \end{tabular}
  \end{center}

As expected, the eigenspace method results in strong improvements over
a Euclidean Distance classifier.  The view similarity approach
with a MLP model of $P(S|B,T_\omega)$ and five hidden units, results
in addition decrease of the error rate of nearly six percent, showing
not only that significant generalization has taken place between
different object models, but that even given a very small training set
of 70 objects, the method actually outperforms an established approach
to object recognition.\footnote{Of course, even better
performance can be achieved by hardcoding additional prior knowledge
about shape and object similarity into the recognition method (e.g.,
\cite{belongie01shape}).  Achieving competitive performance with such
methods would either require encoding additional prior knowledge about
shape similarity in the numerical model of the view similarity function,
or simply using a much larger training set to allow the observer to
learn those regularities directly.}

\section{Discussion}

This paper has introduced the notions of view generalization and view
similarity functions.  We have seen that knowledge of these functions
allows an observer to recognize novel objects from a set of training
view in a Bayes optimal (minimum classification error) way.

By expressing eigenspace and linear combination of view methods in the
framework of view generalization functions, the paper has demonstrated
that fast and compact view generalization functions exist that are at
least as good as commonly used methods for object recognition.
Furthermore, the paper has given a procedure for constructing the
Bayes optimal blurring for matching, a Bayesian version of the
geometric blur method in \cite{berg01}, and shown such blurring
methods to be first order approximations to the view generalization
function.

The paper also reported experiments on the recognition of simulated 3D
paperclips, as well as the recognition of real objects from the
COIL-100 image database of real 3D objects.  In the case of
paperclips, a set of 200 training objects sufficed to reduce the error
rate on single view generalization severalfold compared to 2D view
similarity.  And in the case of the COIL-100 database, the use of view
similarity cut the recognition error rate in half compared to image
based similarity.  This is also one of the first demonstrations of learning
single view 3D generalization for novel objects without requiring
membership in a special object class.

Both the theoretical arguments and the experiments presented in this
paper were only designed to showed that view generalization approaches
are feasible.  We would have expected learning of view generalization
functions to require a large number of training objects.  But
experimental results surpassed expectations and show that view
generalization and view similarity functions that can show significant
amounts of generalization (and actually outperform eigenspace methods)
to arbitrary previously unseen objects are learnable from very modest
numbers of training examples (70 and 200).

Future work has to address a number of practical and engineering
issues.

The experiments in this paper demonstrated single-view generalization.
This was perhaps the more interesting case to address first since few
other methods for 3D object recognition are even capable of performing
meaningful 3D generalization from a single view of an unknown 3D
object.  The extension of this to multi-view generalization requires
some additional tricks; in particular, instead of learning
$P(S=1|B,T_{\omega,1},\ldots,T_{\omega,r})$, it turns out to be
desirable instead to learn $P(S=1|B,f(T_{\omega,1},\ldots,T_{\omega,r}))$
for a function $f$ that ``summarizes'' the views in a way that makes
it easier to learn the view similarity function.

The statistical models used in the experiments in this paper
(empirical distributions and multilayer perceptrons) incorporated no
prior knowledge about objects or shape similarity.  Work on
appearance-based 3D object recognition under 2D transformations (e.g.,
\cite{belongie01shape}, among many others) show that 
systems based on hardcoding knowledge about transformations and shape
similarity into view similarity measures can by themselves achieve a
significant ability to generalize across different 3D views.  Such
techniques can be combined with the adaptive view generalization
approaches presented in this paper.  If such hybrid systems are
constructed carefully, they will perform no worse than the underlying
systems using hardcoded similarity measures, but have the potential
to improve their performance adaptively.  Demonstrating this
also remains for a future paper.

And while it is interesting that view similarity and view
generalization methods can already learn some generalization from as
few as 70 images, training on much larger datasets is clearly
desirable.  After all, we are trying to approximate a similarity
measure that performs Bayes-optimal recognition over the entire
distribution of possible 3D shapes.  Fortunately, it is easy to
generate large amounts of training data without manual labeling from
video sequences, by taking advantage of the fact that video is often
composed of scenes within which individual objects undergo motion
relative to the camera; frames from such scenes provide training
samples for $P(S=1|B,T_\omega)$, while frames from different scenes
can be used as training samples for $P(S=0|B,T_\omega)$.

\bibliographystyle{plain}
\bibliography{iccv}

\end{document}